\documentclass{article}

\usepackage[dblblindworkshop, final]{neurips_2025}

\usepackage[utf8]{inputenc} % allow utf-8 input
\usepackage[T1]{fontenc}    % use 8-bit T1 fonts
\usepackage{hyperref}       % hyperlinks
\usepackage{url}            % simple URL typesetting
\usepackage{booktabs}       % professional-quality tables
\usepackage{amsfonts}       % blackboard math symbols
\usepackage{nicefrac}       % compact symbols for 1/2, etc.
\usepackage{microtype}      % microtypography
\usepackage{xcolor}         % colors
\usepackage{amsmath}
\usepackage{graphicx}
\usepackage{subcaption}

\title{Towards Uncertainty Quantification in Generative Model Learning}

\author{%
  Giorgio Morales, Frederic Jurie, Jalal Fadili \\
  Normandie Univ, UNICAEN, ENSICAEN, CNRS, GREYC \\
  14000, Caen, France \\
  \texttt{giorgiomorales@ieee.org, frederic.jurie@unicaen.fr, jalal.fadili@ensicaen.fr}
}

\begin{document}

\maketitle

\begin{abstract}
  While generative models have become increasingly prevalent across various domains, fundamental concerns regarding their reliability persist.
A crucial yet understudied aspect of these models is the uncertainty quantification surrounding their distribution approximation capabilities. 
Current evaluation methodologies focus predominantly on measuring the closeness between the learned and the target distributions, neglecting the inherent uncertainty in these measurements.
In this position paper, we formalize the problem of uncertainty quantification in generative model learning. 
We discuss potential research directions, including the use of ensemble-based precision-recall curves.
Our preliminary experiments on synthetic datasets demonstrate the effectiveness of aggregated precision-recall curves in capturing model approximation uncertainty, enabling systematic comparison among different model architectures based on their uncertainty characteristics.
\end{abstract}

\section{Introduction}

The use of generative artificial intelligence (GenAI) has significantly impacted applications that are integral to everyday life for a diverse range of users, including mobile applications, content generation tools, and search engines~\cite{GenAI}.
As a result, these models are no longer confined to small teams of researchers or expert users in highly specialized domains.
However, they often exhibit complex, opaque behaviors that are difficult for humans to interpret. 
As these models are rapidly deployed in high-stakes domains, enhancing their reliability and trustworthiness is no longer optional, but a critical necessity.
Therefore, it is essential to quantify their uncertainty to establish a measure of confidence in their learned distributions and mitigate potential risks associated with their deployment~\cite{trustgenai}.

Existing works on uncertainty quantification (UQ) primarily address \textit{sample-level} uncertainty; e.g., the confidence individual outputs (see Appendix~\ref{sec:related} for a review)~\cite{adaptiveUQ,lookbeforeyouleap,ekmekci2023quantifying,aleatoricVSepistemic,deepUQ,Zhang_2024_WACV}.
However, these methods overlook the uncertainty of the \textit{evaluation metric itself}. 
To our knowledge, no work quantifies the confidence in the measured closeness between the learned and target \textit{distributions}.
Addressing this gap would enable more robust evaluations and comparative studies to assess model stability, improving both training methodologies and uncertainty-aware decision-making.

This aspect is crucial for assessing model reliability, especially in applications that rely on precise distribution alignment.
In fields such as experimental particle physics, generative models are increasingly explored as efficient alternatives to traditional Monte Carlo (MC) simulations~\cite{HASHEMI2024100092,kita2024generative}.
MC simulations are computationally expensive yet fundamental for comparing experimental data with theoretical predictions. 
In such applications, ensuring that the learned distribution closely aligns with the target distribution is critical, as small discrepancies might lead to incorrect interpretations of physical phenomena.
Similar challenges arise in weather forecasting, medicine, and molecular biology, where misaligned distributions can lead to unrealistic climate predictions~\cite{weather}, unreliable diagnostic models~\cite{Tiv_Hallucination_MICCAI2024}, or non-physical molecular structures~\cite{molecular}.

Our contribution is two-fold. 
First, we provide a formal definition of UQ in generative model learning, distinguishing it from existing approaches that quantify uncertainty in generative models. 
Second, we outline potential research directions to improve the understanding and measurement of uncertainty in this context.
To establish a foundation for our analysis, we recognize that uncertainty is generally categorized into aleatoric uncertainty, which arises from inherent noise in the data, and epistemic uncertainty, which reflects a lack of knowledge due to limited data coverage or model uncertainty~\cite{hullermeier_aleatoric_2021,nguyen_how_2022}.
In this position paper, we specifically focus on \textit{model uncertainty}, which stems from the choice of model architecture, optimization procedure, and initialization variability. 

To begin addressing this, we suggest estimating model uncertainty by analyzing the variability of the precision-recall curve~\cite{prec_rec,sykes2024} across multiple training runs with different random initializations.
By quantifying this variability, we provide insights into how sensitive a generative model is to training instabilities, which is crucial for assessing its reliability in real-world applications.
Preliminary experiments on a synthetic dataset and a diffusion model suggest that ensembles of PR curves help assess how model uncertainty changes with the score network's parameterization, allowing for selecting the most suitable model architecture.
The aim is not to prescribe a particular method but to highlight a gap: current practice does not capture or communicate uncertainty in generative model learning adequately. 
Closing this gap will require methodological innovations and a shift in community norms so that UQ is treated as essential rather than optional in generative modeling.

\section{Uncertainty in Generative Modeling} \label{sec:unc_def}

Let $\mathbf{X} = \{\mathbf{x}_1, \mathbf{x}_2, \dots, \mathbf{x}_N\}$ be a dataset of $N$ i.i.d. samples from an unknown distribution $P_r$ over a sample space $\mathcal{X}$. 
Let $\mathcal{G}$ denote a fixed generator model family, comprising all functions $g : \mathcal{Z} \rightarrow \mathcal{X}$ defined by a specific architecture, hyperparameter configuration, and training procedure, where $\mathcal{Z}$ is the latent space. 
Let $\Theta \subseteq \mathbb{R}^d$ represent the space of possible model initializations, corresponding to the parameter space's dimensionality. For each $\theta \in \Theta$, we denote by $g^{(\theta)} \in \mathcal{G}$ the generator obtained after training on dataset $\mathbf{X}$ with initialization $\theta$.
Each instantiation $g^{(\theta)}$ induces a learned distribution $P_g^{(\theta)} := g^{(\theta)}\# P_z$ by mapping latent samples under the latent prior $P_z$.
% The divergence between $P_r$ and $P_g^{(\theta)}$ is denoted by $\mathbb{D}(P_r, P_g^{(\theta)})$, where $\mathbb{D}(\cdot,\cdot)$ is a discrepancy or divergence metric (e.g., Jensen-Shannon, Wasserstein, etc.).
We denote by $\mathbb{D}(P_r, P_g^{(\theta)})$ a generic metric quantifying the relationship between $P_r$ and $P_g^{(\theta)}$, which may capture notions of similarity or discrepancy (e.g., Jensen-Shannon divergence, Wasserstein distance).
Its interpretation, whether larger or smaller values are preferable, depends on the specific metric employed.

Let the model initialization $\theta$ be drawn from a distribution $\mu$ over the initialization space.
The model-induced evaluation uncertainty quantifies the variability in the metric $\mathbb{D}(P_r, P_g^{(\theta)})$ due to differences in model initialization (i.e., the randomness in training dynamics).
It is defined as the variance under $\mu$: $
\mathcal{U}_{model} = \mathbb{E}_{\theta \sim \mu}[\mathbb{D}(P_r, P_g^{(\theta)})^2] - \mathbb{E}_{\theta \sim \mu}[\mathbb{D}(P_r, P_g^{(\theta)})]^2.$
This uncertainty reflects the representational and inductive biases of the model family $\mathcal{G}$, which constrain the set of distributions that can be learned and influence which solutions are preferred during training.
For example, if $\mathcal{G}$ is highly expressive and training dynamics are sensitive to initialization, especially under limited training data, the outcomes may vary significantly, increasing $\mathcal{U}_{model}$.

Since $P_r$ and $P_g^{(\theta)}$ are not available in closed form, we estimate $\mathbb{D}(P_r, P_g^{(\theta)})$ using empirical distributions.
The model $g^{(\theta)}$ is used to generate a finite set of $M$ samples $\mathbf{X}_g^{(\theta)} = \{\mathbf{x}'_1, \mathbf{x}'_2, \dots, \mathbf{x}'_{M}\}$.
Given $\mathbf{X} \sim P_r$ and $\mathbf{X}_g^{(\theta)} \sim P_g^{(\theta)}$, the empirical measures are calculated as $\hat{P}_r = \frac{1}{N} \sum_{i=1}^N \delta_{\mathbf{x}_i} \text{  and  } \hat{P}_g^{(\theta)} = \frac{1}{M} \sum_{i=1}^M \delta_{\mathbf{x}'_i},$
where $\delta_{\mathbf{x}'_i}$ is the Dirac measure on ${\mathbf{x}'_i}$.
Therefore, to explicitly incorporate the variability introduced by using finite sets of real samples ($\mathbf{X}$ to estimate $P_r$) and generated samples ($\mathbf{X}_g^{(\theta)}$ to estimate $P_g^{(\theta)}$) when calculating the evaluation metric, 
% account for sampling uncertainty due to finite datasets, 
we define the total evaluation uncertainty as:
\begin{equation}\vspace{-0.5em}
    \label{eq:utotal}
    \mathcal{U}_{total} = \mathbb{E}_{\theta \sim \mu}[\mathbb{E}_{\mathbf{X}, \mathbf{X}_g^{(\theta)}} [\mathbb{D}(\hat{P}_r, \hat{P}_g^{(\theta)})^2 ] ] - \mathbb{E}_{\theta \sim \mu}[ \mathbb{E}_{\mathbf{X}, \mathbf{X}_g^{(\theta)}} [ \mathbb{D}(\hat{P}_r, \hat{P}_g^{(\theta)}) ] ]^2.
\end{equation}
\vspace{-1em}

Note that $\mathbb{D}(\hat{P}_r, \hat{P}_g^{(\theta)})$ can be bounded using Probably Approximately Correct (PAC)-Bayes theory~\cite{pacloss,pac} or other information-theoretic tools~\cite{DMVAEbounds,bortoli,husain,DMbounds}.
For instance, Chen \textit{et al.}~\cite{DMbounds} recently derived upper bounds on the KL divergence between the true data distribution and the distribution induced by a diffusion model, proving that a trade-off exists between generalization performance and the diffusion time used during training.
In general, consider that we obtain an upper bound $B(\theta)$ such that $\mathbb{D}(\hat{P}_r, \hat{P}_g^{(\theta)}) \leq B(\theta)$.
Then, $\mathbb{E}_{\mathbf{X}, \mathbf{X}_g^{(\theta)}}[\mathbb{D}(\hat{P}_r, \hat{P}_g^{(\theta)})^2] \leq \mathbb{E}_{\mathbf{X}, \mathbf{X}_g^{(\theta)}}[B(\theta)^2]$.
As such, the total uncertainty can be bounded as $
    \mathcal{U}_{total} \leq \mathbb{E}_{\mathbf{X}, \mathbf{X}_g^{(\theta)}}[B(\theta)^2] - \mathbb{E}_{\theta \sim \mu}[ \mathbb{E}_{\mathbf{X}, \mathbf{X}_g^{(\theta)}} [ \mathbb{D}(\hat{P}_r, \hat{P}_g^{(\theta)}) ] ]^2.$

However, an upper bound on $\mathbb{D}(\hat{P}_r, \hat{P}_g^{(\theta)})$ is not a proper measure of uncertainty in itself. Such bounds describe worst-case scenarios that are often overly conservative and, crucially, do not quantify the variability or confidence in the learned model. 
% This makes them ill-suited for the practical goal of comparing the stability of different model architectures, which requires understanding typical, not just maximum, variability.
% Similarly, a bound on the total uncertainty $\mathcal{U}_{total}$ does not reflect the actual spread or structure of the evaluation metric values across initializations and datasets; instead, it only limits the maximum possible variability.
The same limitation applies to a bound on the total uncertainty $\mathcal{U}_{total}$, which constrains the maximum possible variability but provides no insight into the distribution of the evaluation metric $\mathbb{D}$ values across initializations or datasets.
For practical applications such as model selection, risk estimation, or active data acquisition, it is necessary to develop and use uncertainty metrics that quantify model confidence and its dependence on both initialization and finite data, rather than relying solely on upper bounds.

\section{Precision-Recall Curve Aggregation} \label{sec:PRcurves}

In this section, we use the concept of precision-recall (PR) curves as they are one of the most commonly used tools for evaluating the performance of generative models~\cite{prec_rec}.
In general, a PR pair $(\alpha, \beta) \in \mathbb{R}^+ \times \mathbb{R}^+$ is defined by the existence of a probe distribution $\mu \in \mathcal{M}_p(\Omega)$, where $\mathcal{M}_p(\Omega)$ is the set of probability distributions over the measurable space $\Omega$, such that $P_r, P_g \in M_p(\Omega)$ and $ P_r \geq \beta \mu $ and $ P_g \geq \alpha \mu $. 
This ensures that $\mu$ extracts a proportion $\beta$ of the real data distribution $P_r$ and $\alpha$ of the generated distribution $P_g$~\cite{prec-rec2,sykes2024}.

Adopting the framework proposed by Simon \textit{et al.}~\cite{prec-rec2}, the PR curve for generative models is obtained through a two-sample classification problem.
Let $U \sim \text{Bernoulli(1/2)}$ be a random variable that determines whether a sample $Z$ is drawn from $P_r$ or $P_g$.
Hence, we define $Z = UX + (1 - U) Y \sim \frac{1}{2} (P_r + P_g)$, where $X \sim P_r$ and $Y \sim P_g$.
The classification task consists of predicting whether $Z = X \sim P_r$ (i.e., $U = 1$) based on a classifier $f \in \mathcal{F}$ from a hypothesis class $\mathcal{F}$.

The PR curve is given by $\partial \text{PR}(P_r, P_g)= \{ (\alpha_{\lambda}, \beta_{\lambda}) \,|\, \lambda \in \mathbb{R}^+ \}$, where each point on the curve corresponds to a different trade-off between false positives and false negatives, controlled by the parameter $\lambda$.
In practice, the curve is computed as a discrete set $\partial \text{PR}(P_r, P_g)= \{ (\alpha_{\lambda}, \beta_{\lambda}) \,|\, \lambda \in \Lambda \}$, where $\Lambda = \{\lambda_k\}_{k=1}^{N_{\Phi}} = \{\tan(\phi_k)\}_{k=1}^{N_{\Phi}}$ denotes the set of tested slope values and $\phi_k \in [0, \frac{\pi}{2}]$.
The false positive rate and false negative rate of $f$ are $\text{fpr}(f) = P[f(X) < \lambda \,|\,  X \sim P_r]$ and $\text{fnr}(f) = P[f(Y) < \lambda \,|\,  Y \sim P_g]$.
The functions $\alpha_{\lambda}$ and $\beta_{\lambda}$ are then defined as $
 \alpha_{\lambda} (P_r,P_g) = \min_{f \in \mathcal{F}} \left\{ \lambda \cdot \text{fpr}(f) +\text{fnr}(f) \right\} \text{and } \beta_{\lambda} (P_r,P_g) = \min_{f \in \mathcal{F}} \{ \text{fpr}(f) +\frac{\text{fnr}(f)}{\lambda} \}.$
% These values are derived from a set of linearly spaced angles $\Theta = \{\theta_k\}_{k=1}^{N_{\Phi}}$ between $\varepsilon$ and $\frac{\pi}{2} - \varepsilon$, where $\varepsilon$ is a small number (e.g., 1e-10), and mapped via $\lambda_k = \tan(\theta_k)$, including boundary values $\lambda_0 = 0$ and $\lambda_{N_{\Phi}} = \infty$.
To quantify the uncertainty in a generative model’s ability to approximate the real distribution, we propose using ensembles of PR curves.
Each curve corresponds to a model trained with a different initialization, and aims to provide insight into how sensitive the generative model is to variations in training conditions.
By analyzing the spread of these curves, an estimate of the model's uncertainty is obtained.

As discussed in Section~\ref{sec:unc_def}, any metric comparing the true and generated distributions is computed using their empirical counterparts.
Given an ensemble of $M$ models with different initializations $\{\theta_1, \dots, \theta_M\}$, let $\partial \text{PR}(\hat{P}_r, \hat{P}^{(\theta_m)}_g)$ represent the PR curve obtained from the $m$-th model.
Since all curves are computed using the same set $\Lambda$, UQ bounds could be constructed along each $\lambda_k$.
This approach would lead to plots with radial dispersion, making interpretation challenging (see Appendix~\ref{sec:appB}).
Instead, we measure dispersion along the precision axis, providing a clearer visualization of variability.
Hence, precision values are compared across models at the same recall levels. 
However, the values $\beta_{\lambda}$ obtained for each $\lambda_k$ may differ across models, making direct comparisons difficult.

To enable a consistent comparison across models, we define a common recall grid, denoted as $\mathcal{B} = \{\beta_j\}_{j=1}^{N_{\Phi}}$, where $\beta_j$ are linearly spaced recall values in $[0,1]$. 
Each curve $\partial \text{PR}(\hat{P}_r, \hat{P}^{(\theta_m)}_g)$ in the ensemble is then interpolated onto this grid, yielding a set of precision values $\alpha_m(\beta_j)$ for each model $m$ at recall $\beta_j$.
Then, the interpolated curve of model $m$ can then be reinterpreted as a multidimensional evaluation metric, $\mathbb{D}(\hat{P}_r, \hat{P}^{(\theta_m)}_g) = \{ \mathbb{D}_{\beta_j}(\hat{P}_r, \hat{P}^{(\theta_m)}_g) = \left( \beta_j, \alpha_m(\beta_j) \right) \}_{j=1}^{N_{\Phi}}$, where each element is a pair representing the model’s precision at recall $\beta_j$. 
%  explicitly state how the multi-dimensional PR curve serves as the specific instantiation of the generic evaluation metric 
This multi-dimensional representation serves as the concrete instantiation of the generic evaluation metric $\mathbb{D}(P_r, P_g^{(\theta)})$ introduced in Section~\ref{sec:unc_def} for the purpose of quantifying evaluation uncertainty in this work.

% at recall $\beta_j$ as $\mathbb{D}_{\beta_j}(\hat{P}_r, \hat{P}^{(\theta_m)}_g) = (\beta_j, \alpha_m(\beta_j))$ $\forall \beta_j \in \mathcal{B}$.
% As such, the entire interpolated PR curve of model $m$ becomes $\mathbb{D}(\hat{P}_r, \hat{P}^{(\theta_m)}_g) = \left\{ \mathbb{D}_{\beta_j}(\hat{P}_r, \hat{P}^{(\theta_m)}_g) \right\}_{j=1}^{N_{\Phi}}$.

For each $\beta_j \in \mathcal{B}$, we compute the mean precision to get the expected value $\mathbb{E}_{\theta \sim \mu}[\mathbb{D}_{\beta_j}(\hat{P}_r, \hat{P}^{(\theta_m)}_g)]$.
A straightforward approach to report the uncertainty at $\beta_j$ is to construct symmetric bounds around this expectation using the standard deviation, as described in Eq.~\ref{eq:utotal}.
However, such symmetric intervals may be misleading when the distribution of precision values is skewed or contains outliers, especially when the ensemble size $M$ is small.
For example, abnormally low precision from a few models at specific recall levels may inflate the estimated uncertainty, potentially pushing upper bounds beyond the valid precision range $[0, 1]$.
To obtain a more robust measure of uncertainty at each $\beta_j$, we instead define the uncertainty interval as the range between the 10th and 90th percentiles of the precision values at $\beta_j$: $
\mathcal{U}_{\beta_j} = \text{Quantile}_{90}\{ \mathbb{D}_{\beta_j}(\hat{P}_r, \hat{P}^{(\theta_m)}_g)\}_{m=1}^M - \text{Quantile}_{10}\{ \mathbb{D}_{\beta_j}(\hat{P}_r, \hat{P}^{(\theta_m)}_g) \}_{m=1}^M.$
This choice provides a robust estimate of the central 80\% performance spread, which is less susceptible to extreme outliers and skewed distributions than symmetric, standard deviation-based intervals.
% This ensures that the range reflects the typical spread of precision values while reducing the influence of extreme outliers.
Finally, the total uncertainty is given by $\mathcal{U}_{total} = \{ \mathcal{U}_{\beta_j} \}_{j=1}^{N_{\Phi}}$.

% Hence, the aggregated PR curve is represented by the mean precision at each recall level,
% \[
% \bar{\alpha}(\beta_j) = \frac{1}{M} \sum_{m=1}^{M} \alpha_m(\beta_j).
% \]

% A straightforward approach to report uncertainty levels at $\beta_j$ would be to use symmetric bounds around the mean, such as $\bar{\alpha}(\beta_j) \pm \sigma_{\alpha}(\beta_j)$, where $\sigma_{\alpha}(\beta_j)$ is the corresponding standard deviation.
% However, this can lead to misleading visualizations, particularly when the distribution of precision values is skewed or contains outliers. 
% For instance, if some models exhibit abnormally low precision at certain recall levels, the standard deviation may be inflated, causing the upper bound to exceed 1, which is not valid for precision values.
% To mitigate this issue, we use the 10th and 90th percentiles of $\{ \alpha_m(\beta_j) \}_{m=1}^M$
% to define uncertainty bounds. 
% This ensures that the range reflects the typical spread of precision values while reducing the influence of extreme outliers.

\section{Preliminary Experiments and Discussion} \label{sec:exp}

The experiments presented in this section represent a \textbf{proof of concept for the ensemble PR curve diagnostic introduced in Section~\ref{sec:PRcurves}} rather than a definitive novel evaluation method.
We used a synthetic dataset generated from a truncated Gaussian ring distribution for evaluation. 
This dataset consists of 20,000 samples arranged in eight clusters uniformly distributed along a circular pattern with a fixed radius of 5. %, as depicted in Fig.~\ref{fig:dataset}. 
Each Gaussian component has a distinct standard deviation, introducing variability in the spread of clusters while maintaining a structured ring formation\footnote{Reproducible code: \url{https://github.com/GiorgioMorales/UQ-PRD}}.
In addition, points were truncated based on their Mahalanobis distance, retaining only samples within approximately three standard deviations from their respective cluster centers.
The dataset was split evenly, with 10,000 samples for training and 10,000 for validation.

We trained standard Denoising Diffusion Probabilistic Models (DDPMs)~\cite{ho_denoising_2020} to approximate our problem's distribution.
Note, however, that the central argument presented in this work is agnostic to the specific generative model employed, as the analysis is conducted post hoc and relies solely on the characteristics of the generated samples.
In this example, we use a residual feedforward neural network with time-dependent embeddings for the score prediction network.
Its first layer projects the two-dimensional inputs into a space of 128 dimensions.
The core consists of $C$ residual blocks, where each transforms its input $h_i$ using two hidden layers with 256 and 128 units, respectively, such that $h_{i+1} = h_i + W_{i,2} (\sigma(W_{i,1} h_i + b_1 + W_e t_e)) + b_2$.
Here, $ t_e$ is the time embedding, matrices $W$ and $b$ are learnable projections, and $\sigma$ is the SiLU activation. The final 
uses two units to project the output back to the original input dimension.
The depth $C$ determines the network’s complexity.
We experimented with $C \in \{1, 2, 4, 8\}$, adjusting the training duration accordingly: 10,000 epochs for $C \in \{1, 2\}$, 15,000 epochs for $C=4$, and 20,000 epochs for $C=8$. 
This choice is based on the assumption that as model complexity increases, so does the time required for convergence. 
However, training for additional epochs beyond these values did not result in noticeable improvements.
For each depth setting, we trained 30 DDPMs.
A sample size of 30 is typically used in hypothesis testing, as it is generally considered sufficient for the central limit theorem to hold.

\begin{figure}[t]
    \centering
\vspace{-2ex}
\includegraphics[width=\linewidth,height=0.18\textheight]{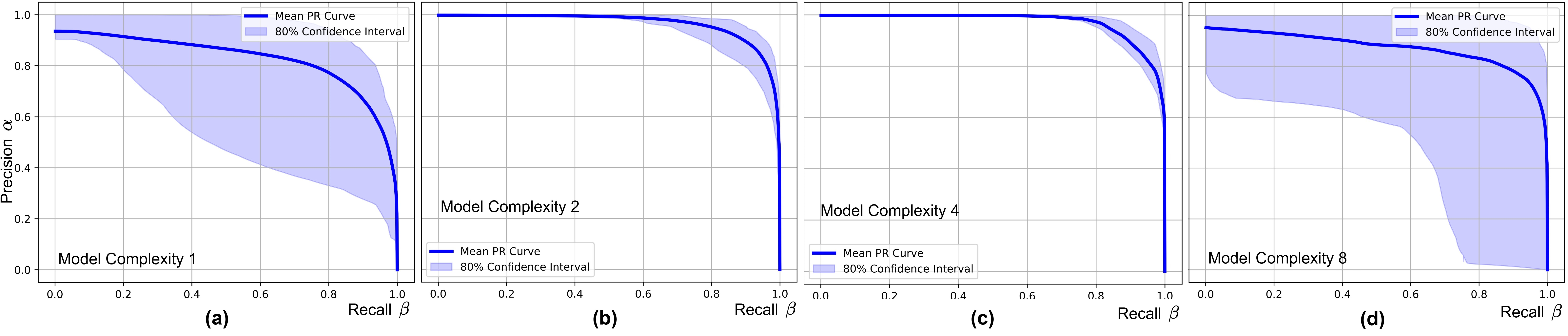}
\vspace{-3ex}
    \caption{PR curves ensembles obtained with $C=$ \textbf{(a)} $1$, \textbf{(b)} $2$, \textbf{(c)} $4$, and \textbf{(d)} $8$.}
    \label{fig:PRComparison}
\vspace{-2ex}
\end{figure}

\begin{figure}[t]
    \centering
    \begin{minipage}[b]{0.64\linewidth}
        \centering
        \includegraphics[width=\linewidth]{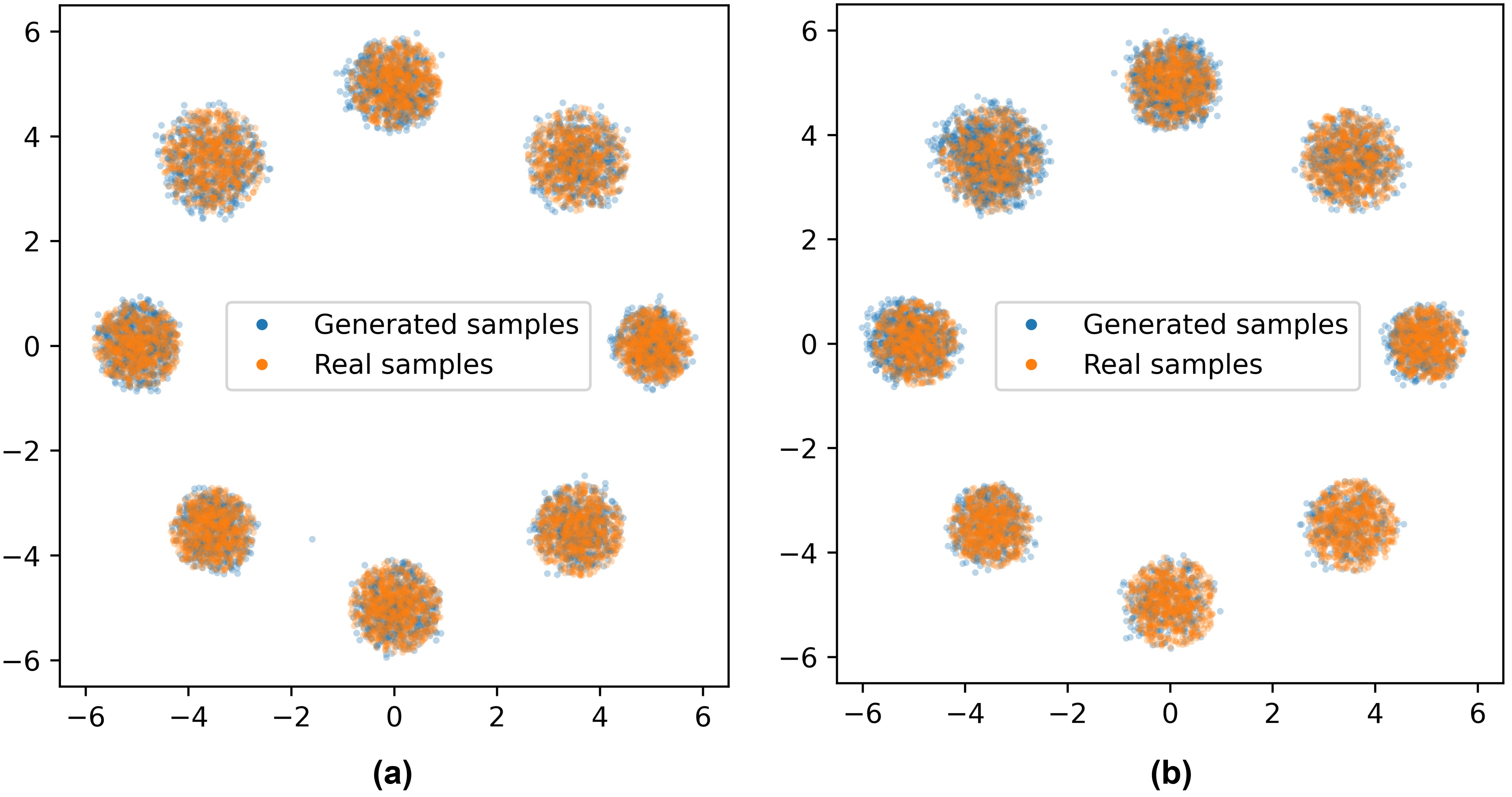}
        \vspace{-2ex}
        \caption{Real vs. generated samples ($C=4$). \textbf{(a)} Best-performing and \textbf{(b)} Worst-performing model.}
        \label{fig:C4}
    \end{minipage}\hfill
    \begin{minipage}[b]{0.33\linewidth}
        \centering
        \includegraphics[width=\linewidth]{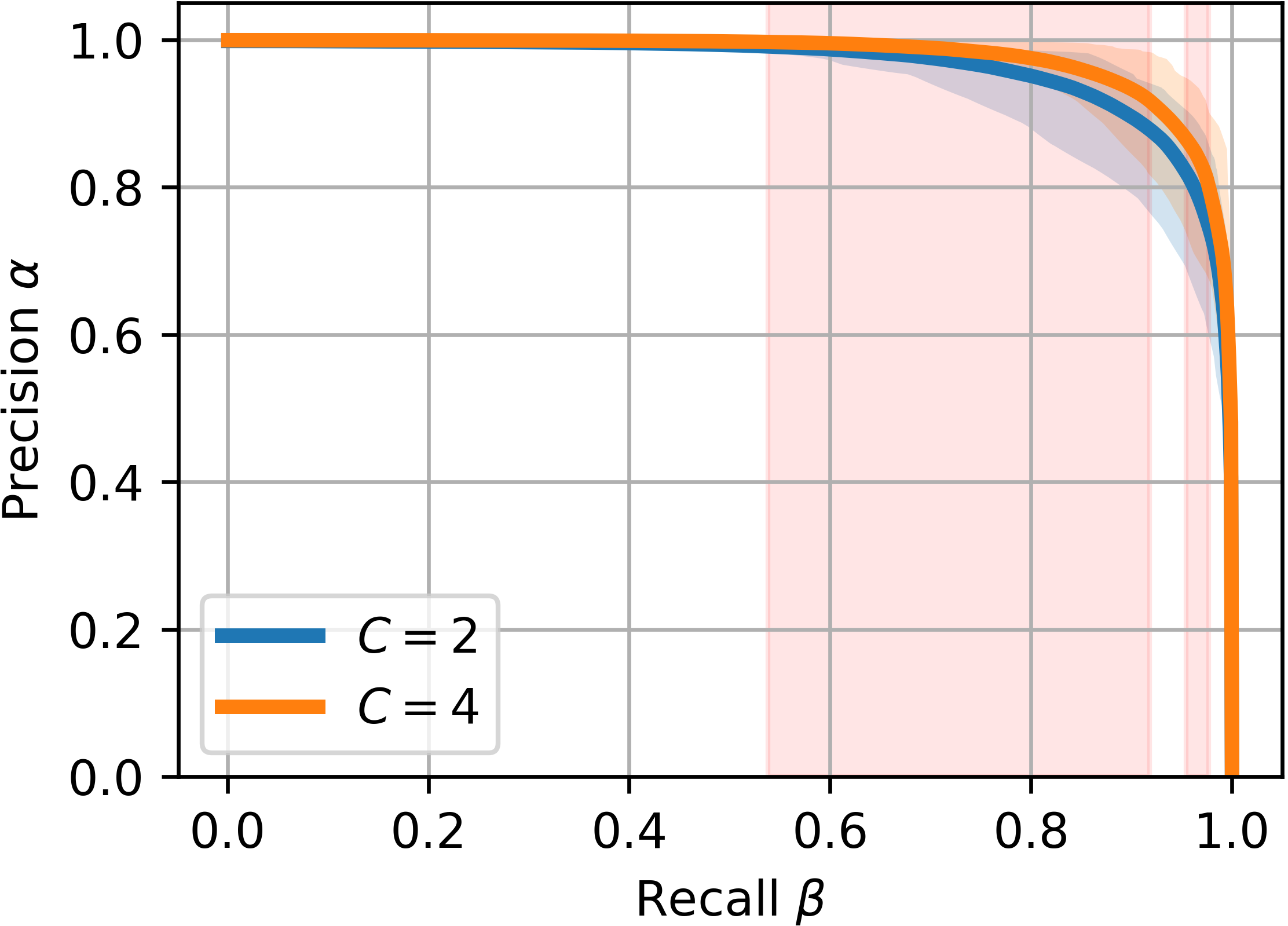}
        \vspace{-2ex}
        \caption{PR curve ensembles ($C=2 \text{ vs. } 4$). Red areas indicate statistical significance.}
        \label{fig:significance}
    \end{minipage}
\end{figure}

We generated PR curves following the $k$-nearest neighbors-based methodology proposed by Sykes \textit{et al.}~\cite{sykes2024}, although our central argument remains independent of the implementation used to compute PR curves.
Fig.~\ref{fig:PRComparison} presents the aggregated PR curves for each model ensemble, depicting the mean PR curve along with the 80\% confidence interval, computed using the 10th and 90th percentiles (Section~\ref{sec:PRcurves}).
These results indicate that using $C=2$ and $C=4$ yielded the narrowest confidence intervals, suggesting the lowest model uncertainty.
For instance, Fig.~\ref{fig:C4} illustrates the original and generated samples produced by the best- and worst-performing models with $C=4$.
Here, the best-performing model is defined as the one with the highest area under the PR curve, while the worst-performing model corresponds to the one with the lowest area.

% \begin{figure}[t]
%     \centering
%     \includegraphics[width=.68\linewidth]{images/C42.jpg}
%     \vspace{-2ex}
%     \caption{Real vs. generated samples ($C=4$). \textbf{(a)} Best-performing and \textbf{(b)} Worst-performing model.}
%     \label{fig:C4}
% \end{figure}

% \begin{figure}[t]
%     \centering
%     \includegraphics[width=0.3\linewidth]{images/significance.png}
%     \vspace{-1ex}
%     \caption{PR curves ensembles ($C=2$ vs $C=4$). Red areas indicate statistical significance.}
%     \label{fig:significance}
% \end{figure}

We argue that the trend in Fig.~\ref{fig:PRComparison} illustrates the relationship between model complexity and stability.
High variance is obtained when $C=1$, producing high-quality PR curves in a few cases, suggesting the model may have sufficient capacity to approximate the target distribution.
Nevertheless, we attribute the observed high variance to an irregular and fragmented loss landscape that increases the likelihood of convergence to suboptimal local.
In contrast, the improved stability of intermediate models ($C=2,4$) is indicative of a smoother or more connected loss surfaces that facilitate more reliable convergence and reduce variance across runs~\cite{losssurface,landscape}.
At the other end of the spectrum, when $C=8$, excessive model complexity may lead to overfitting, making the learned distribution highly sensitive to initialization and training dynamics.  
As a result, both underparameterized and overparameterized models exhibit higher model uncertainty, whereas intermediate levels of complexity achieve a more favorable trade-off between capacity and stability.
% This demonstrates a favorable trade-off between model capacity and training stability at intermediate levels of complexity.

Furthermore, Fig.~\ref{fig:significance} compares the ensemble of PR curves obtained for $C=2$ and $C=4$.
Given that we have a sample of 30 interpolated precision values for each $\beta_j \in \mathcal{B}$, we performed a hypothesis test to compare the distributions obtained for each model type.
Specifically, we conducted a paired $t$-test between $\{ \alpha_m^{(C=2)} (\beta_j) \}_{m=1}^{30}$ and $\{ \alpha_m^{(C=4)} (\beta_j) \}_{m=1}^{30}$ $\forall \beta_j \in \mathcal{B}$.
Fig.~\ref{fig:significance} highlights in red the regions where model $C=4$ significantly outperformed model $C=2$ at the 0.05 significance level.
From this, we conclude that model $C=4$ exhibited significantly better performance than model $C=2$ in 44.2\% of the curve.
It is important to note that our suggestion of generating interpolated precision values based on a common recall set relies on the assumption that $N_{\Phi}=500$ is sufficiently large to produce smooth PR curves, making interpolation errors negligible.
To validate this, we tested $N_{\Phi}=\{250, 500, 1000\}$ and observed no noticeable differences in the shape of the PR curves.

To further investigate the role of epistemic uncertainty, we conducted another experiment focused on evaluating the variability introduced by finite data availability. 
Note that, in our setup, aleatoric uncertainty corresponds to the inherent ambiguity in the data itself, such as the natural overlap between the Gaussian clusters.
While our previous analysis fixed the dataset size and varied model complexity to assess model-induced uncertainty, this experiment isolates the effect of limited training data by fixing the model architecture to $C=4$ and varying the dataset size. 
We used dataset sizes of 20,000, 15,000, 10,000, and 5,000, each split evenly into training and validation sets.
For each size, we trained ensembles of 30 models following the same procedure used in the previous experiment and generated corresponding ensembles of PR curves.  
% Note that this experiment probes epistemic uncertainty (due to limited data); in our setup, aleatoric uncertainty corresponds to the inherent ambiguity in the data itself, such as the natural overlap between the Gaussian clusters.

\begin{figure}[t]
    \centering
    \includegraphics[width=1.0\linewidth]{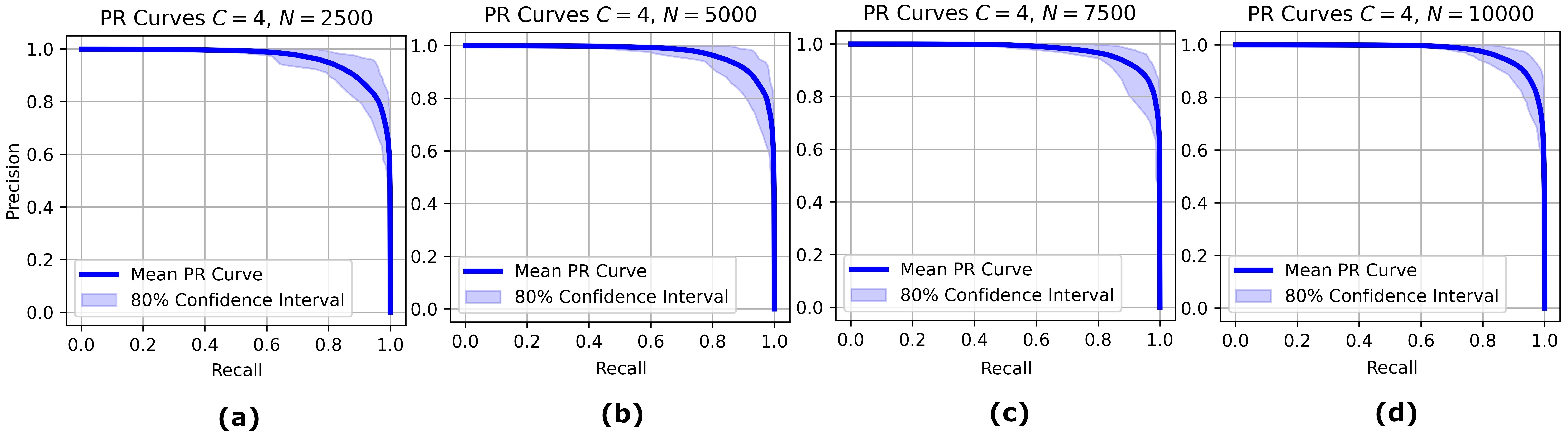}
    % \vspace{-4ex}
    \caption{PR curves obtained for $C=4$ and $N=$ \textbf{(a)} $2500$, \textbf{(b)} $5000$, \textbf{(c)} $7500$, \textbf{(d)} $10000$.}
    \label{fig:datasetsizePR}
\end{figure}

The results shown in Fig.~\ref{fig:datasetsizePR} allow us to analyze how the reduced availability of training samples influences the stability and performance of the learned distributions.
As expected, wider confidence intervals were obtained when using smaller datasets, reflecting the increasing epistemic uncertainty resulting from reduced empirical support.
In addition, smaller datasets also led to worse performance metrics, as the area under the PR curves decreases consistently as the dataset size is reduced and they deviate further from the ideal precision-recall values of $1$.
As such, this experiment highlights the extent to which smaller datasets can degrade reliability, increase model variability, and impact the ability of the generative model to approximate the target distribution accurately.

Nevertheless, some caveats are important.
First, generating ensembles of PR curves to quantify uncertainty can be computationally prohibitive, particularly for complex generative modeling tasks.
However, we argue that in applications where reliability is critical (e.g., scientific simulations), the need for robust and trustworthy models justifies the additional computational cost.
Nevertheless, future work will explore alternative UQ methods, such as Bayesian hypernetworks~\cite{compound}, which can estimate model uncertainty without requiring multiple trained models, thereby reducing computational overhead.
Second, the diagnostic depends on the feature space used to compute distances for the kNN-based estimator and, more generally, on the choice of the PR curve estimator. 
In image or other high-dimensional domains, a learned embedding or pretrained feature extractor is necessary, and feature choice will interact with the measured uncertainty.
Third, interpolating onto a common recall grid and using percentile bands are pragmatic choices designed to visualize dispersion robustly; however, they do not replace formal statistical treatment. 
For example, paired $t$-tests assume approximate normality of paired differences, and multiple testing across recall values requires correction or nonparametric permutation alternatives. 
With $M = 30$, the central limit theorem provides some robustness; nevertheless, bootstrap or permutation methods are safer for small ensemble sizes.

\section{Conclusion}

Generative models are widely used in applications where precise distribution approximation is crucial, yet current evaluation methods overlook uncertainty in these approximations. 
This is concerning in domains like scientific simulations, where minor discrepancies can lead to incorrect conclusions. 
Thus, UQ is essential for assessing reliability and guiding improvements in generative model design.

This work makes two key contributions. 
First, we formally define UQ in generative model learning, highlighting its distinction from existing methods that focus on sample-level rather than distribution-level uncertainty. 
Second, rather than introducing a new algorithm, we suggest ensemble-based precision–recall bands as a simple but effective diagnostic to illustrate why single-curve reporting is insufficient.
Experiments show that PR curve variability across multiple training runs provides meaningful insights into model sensitivity, enabling informed comparisons between architectures.

The central message is that evaluating generative models should not stop at reporting a point estimate of performance. 
Ensemble-based diagnostics are one way to make this uncertainty visible, but they are only a step.
More scalable, theoretically grounded, and domain-aware approaches will be needed if UQ is to become standard in generative modeling.
Future research will explore alternative techniques to estimate generative model uncertainty without requiring multiple training runs. 
Further evaluation on real-world and higher-dimensional datasets will also help refine the proposed methodologies and assess their broader applicability. 
Applying this methodology to high-dimensional data like images might introduce challenges related to the choice of features used for PR curve computation~\cite{prec_rec,sykes2024}, as feature space choice could interact with the measured uncertainty. 
Furthermore, the nature and magnitude of evaluation uncertainty might differ significantly across various model families (e.g., GANs, VAEs, flow-based models) due to their distinct learning dynamics and failure modes.

\begin{acksection}
This research was supported by funding from the ANR under award number ANR-19-CHIA-0017 and the Region Normandie through the DEEPVISION grant.
\end{acksection}

\bibliographystyle{plain}
\small
\bibliography{references}

% \section*{References}

\appendix

% \counterwithin{figure}{section}
% \counterwithin{table}{section}
% \counterwithin{algorithm}{section}
% \counterwithin{definition}{section}
% \counterwithin{equation}{section}

\section{Related Works on Uncertainty Quantification for Generative Models} \label{sec:related}

Uncertainty quantification in generative models can take multiple forms, depending on the specific aspect being measured. 
It may refer to the uncertainty in the generated outputs, the uncertainty of the learned parameters, the uncertainty in the learned distribution itself, or the uncertainty in the estimated closeness between the learned and target distributions. 
Most existing works focus on the first case, aiming to quantify the uncertainty associated with either individual generated samples or sets of samples corresponding to a given class or sub-group.

For example, Ekmekci and Cetin~\cite{ekmekci2023quantifying} introduced an approach that adapts posterior sampling techniques for statistical uncertainty estimation in class-conditioned sample generation.
It employs an ensemble of conditional generative models, where each model processes multiple samples drawn from a prior distribution for a specified target class.
The samples are then modeled as a Gaussian mixture, enabling the calculation of statistical measures that quantify overall uncertainty. 
Building upon Depeweg \textit{et al.}'s framework~\cite{pmlr-v80-depeweg18a}, this uncertainty is further decomposed into its aleatoric and epistemic components.

Ekmekci and Cetin's approach was tailored for medical image generation and assumed that objects corresponding to the same class appear at similar scales and fixed positions, enabling uncertainty quantification at the pixel level. 
However, in more general settings where objects may vary in scale and position, this method would yield high uncertainty estimates consistently due to the increased variability in the generated outputs.
Nevertheless, this type of approach is well-suited for forecasting applications, as demonstrated by Chan \textit{et al.}~\cite{aleatoricVSepistemic}, where the input condition is an image of the current state and the output is a predicted future state.
In this context, pixel-wise uncertainty quantification is more meaningful, as it provides localized confidence estimates for each region of the generated output.
In addition, Chan \textit{et al.} pointed out the computational inefficiency of conventional ensembling methods for complex architectures.
To address this, they introduced Hyper-Diffusion Models, a framework that enables epistemic and aleatoric uncertainty estimation with a single model by learning a hypernetwork that parameterizes an ensemble of diffusion models.

A common approach to quantifying the uncertainty associated with individual model outputs is to measure their divergence.
In that sense, Huang \textit{et al.}~\cite{lookbeforeyouleap} presented a comparative study of eight large language models (LLMs) across three natural language processing (NLP) tasks and a code generation task.
While LLMs are not generative models in the strict sense, their inference process involves a generation mechanism, as they produce token sequences autoregressively based on learned patterns in the training data.
One key finding was that sample-based uncertainty quantification methods provided more informative reports about LLM's uncertain/non-factual predictions.
These methods modify the LLM's temperature parameter to promote diversity in the generated results.
The uncertainty subsequently is quantified using variation ratio (VR) and variation ratio for original prediction (VRO)~\cite{VRVO}, which captures the level of variability in the model’s predictions.
The study highlights a critical limitation: existing uncertainty quantification techniques struggle to detect subtle inconsistencies. 
The authors suggest that more refined distance functions should be developed to improve uncertainty estimation for task-specific evaluations.

Alternatively, the uncertainty of individual outputs can be reported using prediction intervals (PIs), which provide upper and lower bound estimates within which a prediction will fall according to a certain probability~\cite{LUBE}.  
Thus, Kim \textit{et al.}~\cite{adaptiveUQ} presented a conformal prediction method that generates PIs for classification or regression tasks in cases where users do not have access to the training data, as is the case for commercial LLMs.
% According to the $\textit{High-quality (HQ)}$ $\textit{principle}$~\cite{ICML-2018-PearceBZN}, the PIs produced for a given sample should be as narrow as possible while still covering the required proportion of predictions.
To produce narrow but accurate intervals, they use conditioning on subpopulations of data points that share similar characteristics. 
Unlike traditional group-conformal prediction approaches that assume predefined groups, an adaptive partitioning method called Conformal Tree (CT) is proposed to identify such groups dynamically. 
% CT employs tree-based supervised learning to 
It segments the input space into regions (i.e., bins) and applies split-conformal prediction within each region to produce group-specific high-quality PIs.

PIs can also be generated for image generation applications, where upper and lower-bound images provide a pixel-wise confidence metric. %for tasks such as image-to-image generation.
Recently, several PI-generation methods have been developed specifically for diffusion models.
A naïve approach assumes that a trained diffusion model has effectively learned the target distribution. 
This allows multiple samples to be generated and used to compute PI bounds statistically.
Horwitz and Hoshen~\cite{horwitz_conffusion_2022} argued that this approach is computationally inefficient and generates suboptimal bounds.
To address this, they introduced Con\textit{ffusion}, a technique that fine-tunes diffusion models to predict PI images in a single forward pass.
Their approach incorporates a loss function that minimizes the distance between the original model’s generated image and the lower bound, as well as between the upper bound and the generated image.
This method improves computational efficiency compared to sampling-based techniques while producing tighter, more reliable bounds.
In addition, their approach uses a Risk-Controlling Prediction Sets (RCPS) procedure to ensure risk control, providing formal guarantees on the reliability of the generated intervals.
Building on this, Teneggi \textit{et al.}~\cite{teneggi_how_2023} introduced a high-dimensional generalization of RCPS, termed K-RCPS, which formally minimizes the mean interval length while maintaining risk control guarantees.
By leveraging conformal prediction techniques for PI image generation, they offer distribution-free uncertainty guarantees, making it applicable beyond diffusion models.

Furthermore, generative models are widely used for data reconstruction, where uncertainty quantification is essential for assessing reliability.
For instance, Böhm \textit{et al.}~\cite{deepUQ} address Bayesian inverse problems by leveraging a generative model trained on uncorrupted data to reconstruct noisy inputs while dealing with uncertainty through posterior analysis in latent and data space.
Uncertainty is then assessed by analyzing the structure of the learned latent space and visualizing the diversity of reconstructed samples, providing an interpretable representation of reconstruction confidence.
In contrast, Zhang \textit{et al.}~\cite{Zhang_2024_WACV} introduced a deep variational framework that models the posterior distribution of the reconstruction using a flow-based approach, parameterizing the posterior and minimizing its Kullback-Leibler (KL) divergence to an approximate prior. 
To improve stability and expressivity, it utilizes bi-directional regularization, gradient boosting, and a space-filling design for variance reduction in both latent and posterior spaces. 
This method provides pixel-wise uncertainty estimates, allowing fine-grained statistical analysis of reconstructed images.

\section{Radial Dispersion Plots} \label{sec:appB}

\begin{figure}[t]
    \centering
    \begin{subfigure}[t]{0.48\textwidth}
        \centering
        \includegraphics[width=\linewidth]{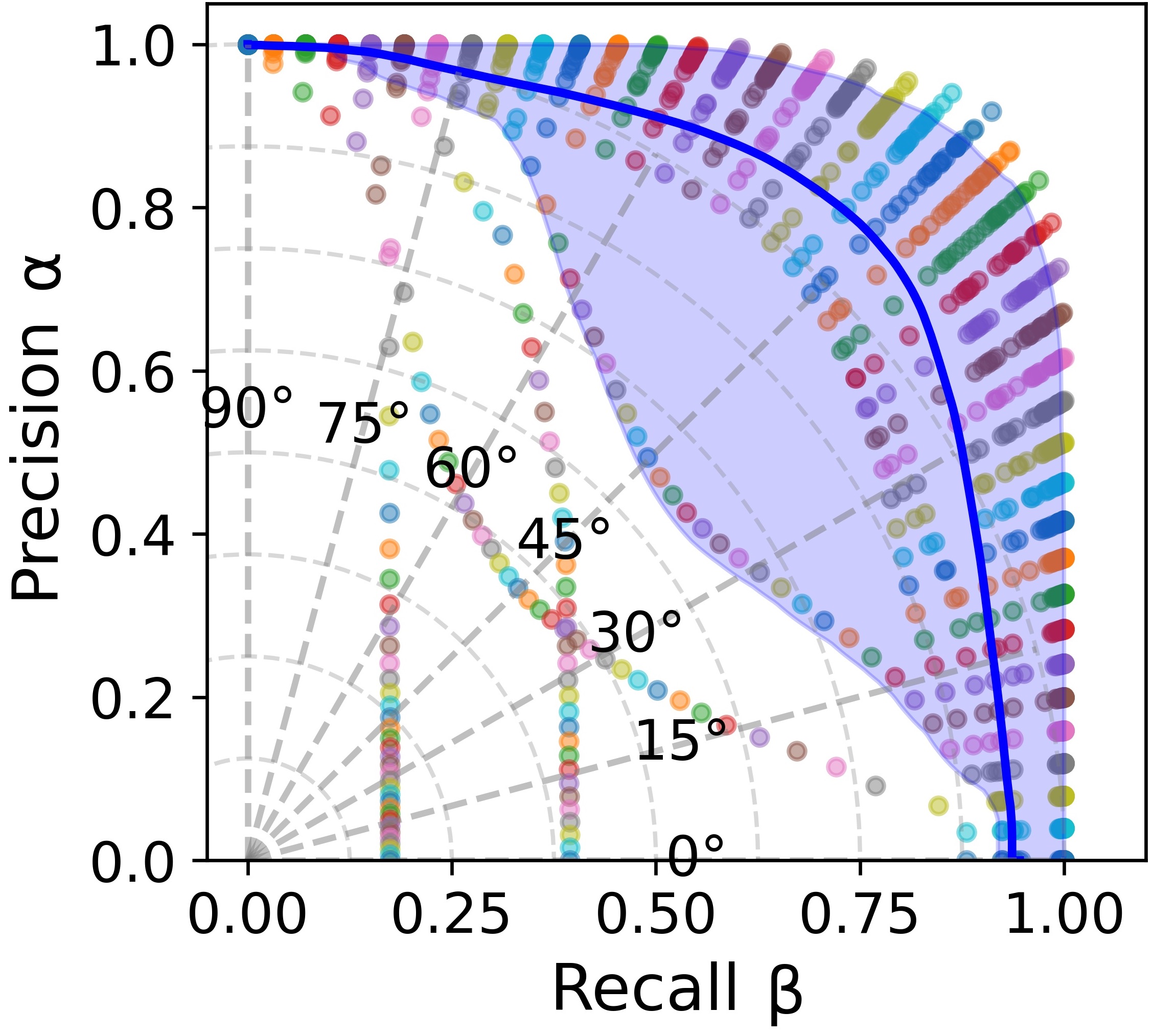}
        \caption{}
    \end{subfigure}
    \hfill
    \begin{subfigure}[t]{0.48\textwidth}
        \centering
        \includegraphics[width=\linewidth]{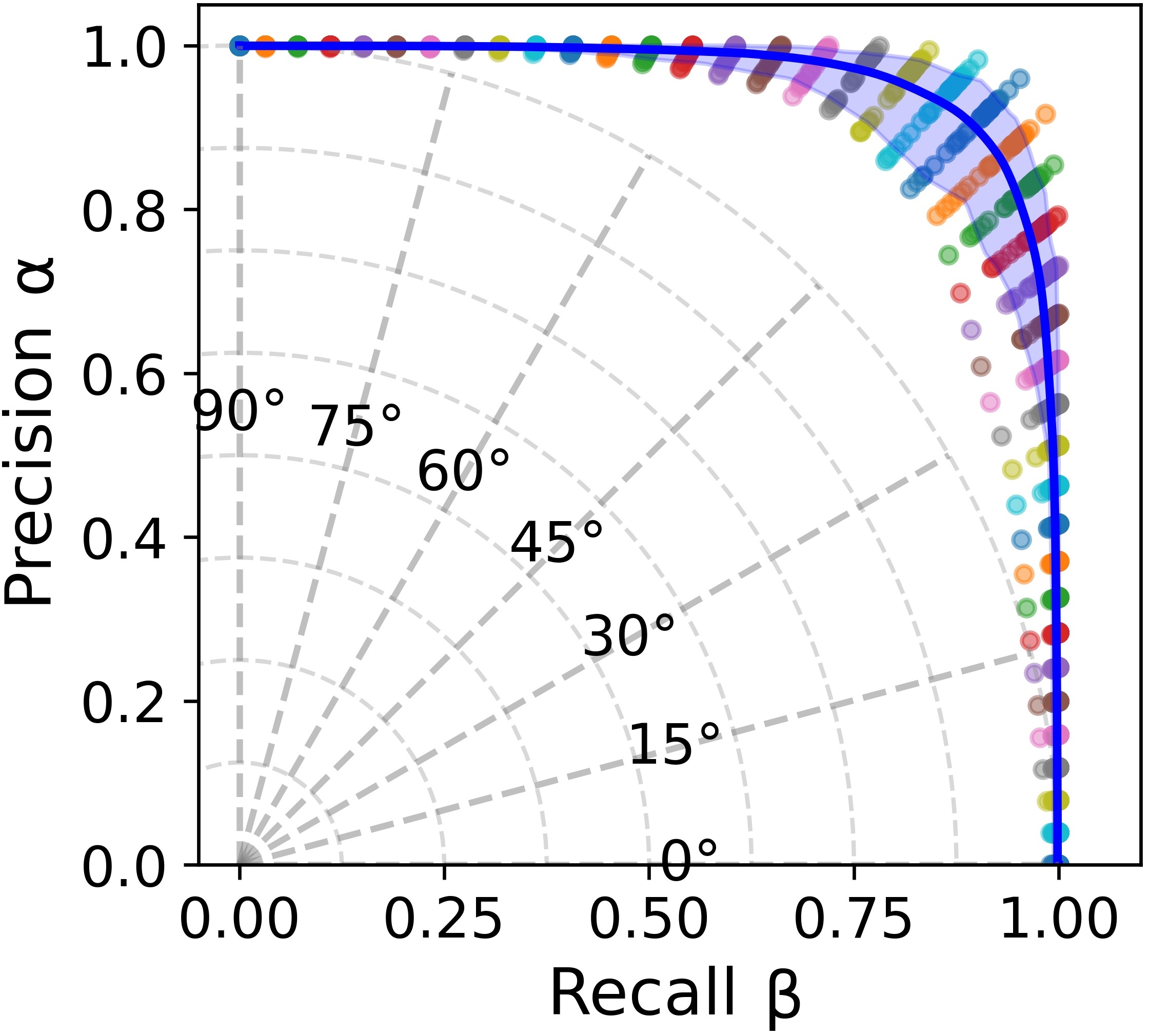}
        \caption{}
    \end{subfigure}
    
    \vspace{0.5em}
    
    \begin{subfigure}[t]{0.48\textwidth}
        \centering
        \includegraphics[width=\linewidth]{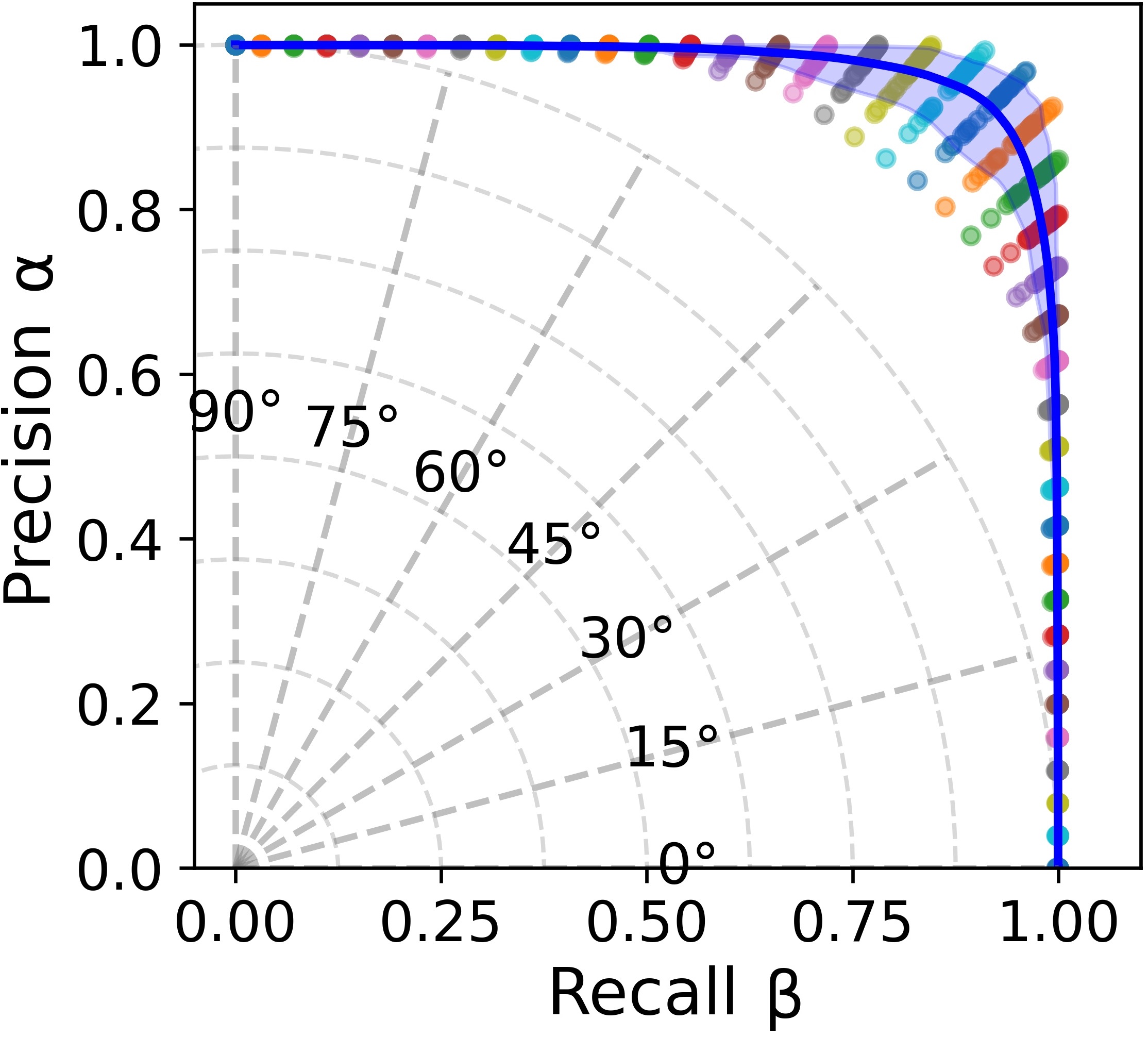}
        \caption{}
    \end{subfigure}
    \hfill
    \begin{subfigure}[t]{0.48\textwidth}
        \centering
        \includegraphics[width=\linewidth]{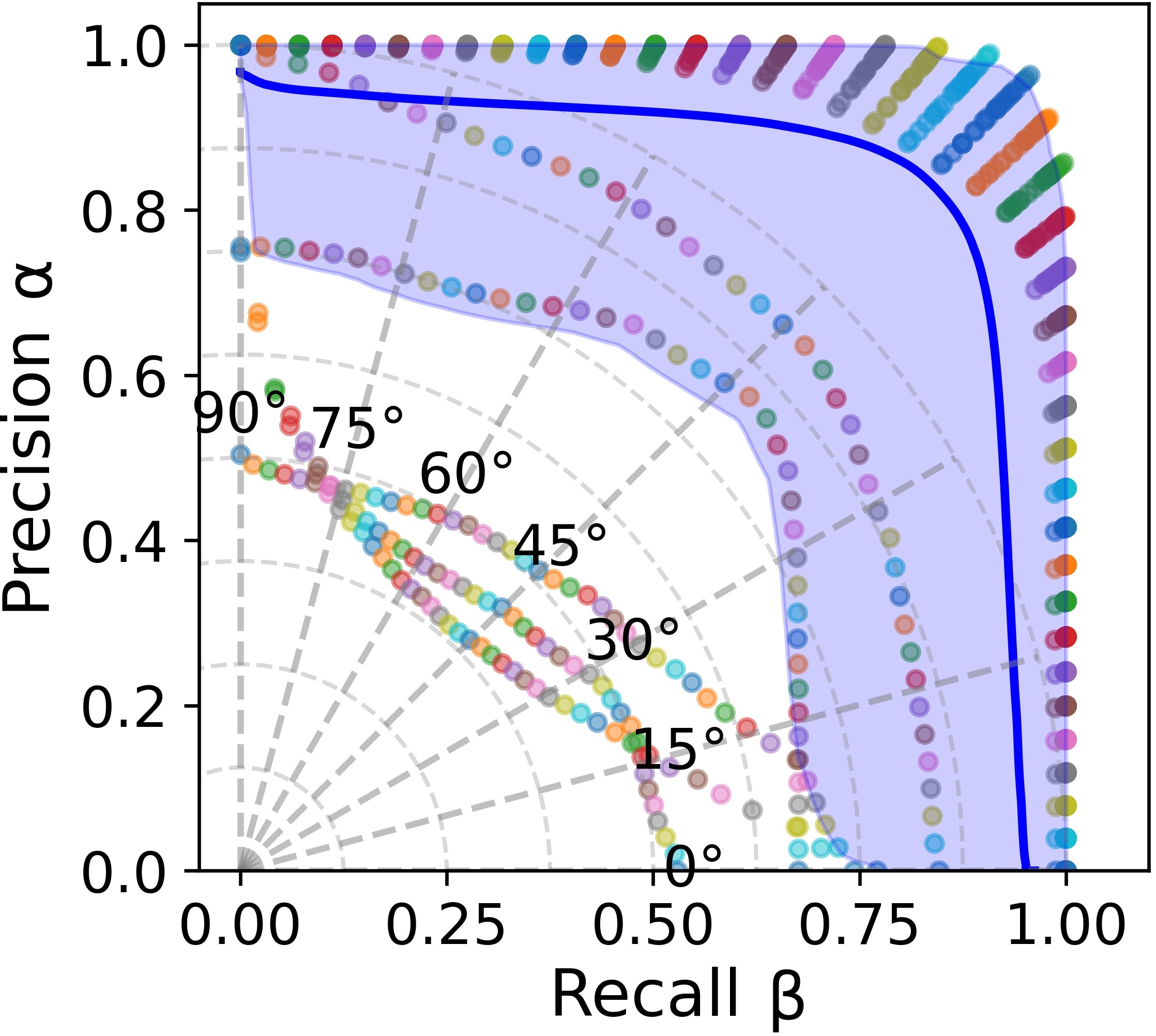}
        \caption{}
    \end{subfigure}

    \caption{Alternative PR curve uncertainty visualization showing radial dispersion at model complexities \textbf{(a)} $C=1$, \textbf{(b)} $C=2$, \textbf{(c)} $C=3$, and \textbf{(d)} $C=4$.}
    \label{fig:lambdak}
\end{figure}

In this section, we present an alternative visualization to that introduced in Section~\ref{sec:PRcurves}.
Recall that all PR curves in $\{ \partial \text{PR}(\hat{P}_r, \hat{P}^{(\theta_1)}_g), \dots, \partial \text{PR}(\hat{P}_r, \hat{P}^{(\theta_M)}_g) \}$ are evaluated over the same set of slopes $\Lambda$.
Consequently, uncertainty bounds can be constructed at each $\lambda_k \in \Lambda$ (equivalently, at each angle $\phi_k \in [0, \pi/2]$).
If curves are compared element-wise, the appropriate direction of comparison is the radial axis, and the analysis naturally focuses on radial dispersion.
Following Section~\ref{sec:PRcurves}, we define the uncertainty interval at slope $\lambda_k$ as the range between the 10th and 90th percentiles of the precision values $\alpha_k$ across the ensemble: $\mathcal{U}_{\lambda_k}
= \operatorname{Quantile}_{90}\!\{ \mathbb{D}_{\lambda_k}(\hat{P}_r, \hat{P}^{(\theta_m)}_g) \}_{m=1}^M
- \operatorname{Quantile}_{10}\!\{ \mathbb{D}_{\lambda_k}(\hat{P}_r, \hat{P}^{(\theta_m)}_g) \}_{m=1}^M,$
where $\mathbb{D}_{\lambda_k}$ denotes the set of precision values $\alpha_k$ obtained at slope $\lambda_k$.
Fig.~\ref{fig:lambdak} illustrates these results such that the shaded region corresponds to the intervals $\mathcal{U}_{\lambda_k}$, while the solid blue curve represents the ensemble mean.
To emphasize the radial interpretation, we additionally display, every five steps along the curve, the $M$ ensemble points associated with each $\lambda_k$, using consistent coloring to indicate angle.
Finally, unlike the interpolated curves shown in Section~\ref{sec:exp}, these plots rely directly on the computed values, which results in slightly different curve shapes.

\end{document}